\documentclass{article}

\RequirePackage{natbib}
\usepackage{graphicx} 
\usepackage[utf8]{inputenc} 
\usepackage[T1]{fontenc}    
\usepackage{hyperref}       
\usepackage{url}            
\usepackage{booktabs}       
\usepackage{amsfonts}       
\usepackage{nicefrac}       
\usepackage{microtype}      
\usepackage{listings}       
\usepackage{mathptmx}       

\usepackage[margin=0.5cm]{caption} 

\newcommand{\specialcellcenter}[2][c]{%
  \begin{tabular}[#1]{@{}c@{}}#2\end{tabular}}

\setcitestyle{authoryear,round,citesep={;},aysep={,},yysep={;}}

\renewcommand{\cite}[1]{\citep{#1}}

\usepackage{blindtext}
\usepackage{geometry}
\begin{document}

{\fontfamily{ptm}\selectfont
\title{Learning the Latent Rules of a Game from Data: \\ A Chess Story}}

\author{Ben Fauber\thanks{Correspondence to: Ben.Fauber@dell.com} \\
\normalsize{Dell Technologies}
}
\date{October 3, 2024}

\maketitle

\begin{abstract}
We demonstrate that small pretrained foundational generative language models with millions of parameters can learn the latent rules of a process from data associated with the process. Inspired by Stefan Zweig's novella \emph{Schachnovelle}, also known as \emph{The Royal Game} in English, we show that 28M and 125M parameter pretrained foundational small language models (SLMs) can be instruction fine-tuned with 1,000-to-1,000,000 examples to learn the rules of chess, propose legal moves, and accurately solve chess problems. We also explore the impact of successive language model fine-tuning epochs on improved outcomes and demonstrate reductions in model hallucinations by increasing the number of instruction fine-tuning examples.
\end{abstract}

\section{Introduction}

The novella \emph{Schachnovelle}, also known as \emph{The Royal Game} in English, was written by the renowned Austrian author Stefan Zweig and first published in 1942 \cite{zweig_royalgame}. The story follows an anonymous narrator on an ocean liner traveling from New York to Buenos Aires. Among the passengers is world chess champion Mirko Czentovic. The narrator and a businessman named McConnor challenge Czentovic to a chess match, in which Czentovic easily wins the first game. In the next game, a mysterious passenger known only as Dr. B. intervenes, confounding Czentovic's strategy and unexpectedly forcing a draw.

As the story progresses, Dr. B. confides in the narrator that he was once a prisoner of the Gestapo. To maintain his sanity during his imprisonment, Dr. B. obsessively studied a book containing 150 championship chess games. The story culminates when Dr. B. agrees to a match against Czentovic. Dr. B. stuns Czentovic by winning the first game but grows increasingly agitated by the slow tempo of the second game. The narrator intervenes before Dr. B.'s frustration with the game’s tempo overwhelms him, prompting Dr. B. to resign and withdraw from the second game as the story concludes.

We have shown that instruction fine-tuning small pretrained generative language models (SLMs) leads to highly specialized models capable of accurately performing challenging tasks that the base models cannot perform \cite{Fauber2024AccuratePO, Fauber2024PretrainedGL}. Our advances in fine-tuning language models prompted us to revisit Stefan Zweig's fictional character, Dr. B., and his process of mastering chess.

We found it compelling that Dr. B., who initially had a limited knowledge of chess, mastered the game after intensely studying a book of championship chess games during his imprisonment. Even more remarkable was that his book contained no board diagrams and only used standard algebraic chess notation (SAN) to chronicle each game, which Dr. B. initially did not understand at all. Dr. B. described the book's contents as, "...what were to me the almost unintelligible symbols... It all seemed a kind of algebra to me, to which I could find no key. Only gradually did I puzzle out that the numbers stood for the ranks and the letters for the files, so that you could establish the position of each piece."

Dr. B.'s fictitious yet plausible experience led us to explore the real possibility of learning the latent rules of a game from data. Herein, we explore: 1) can an instruction fine-tuned language model learn the rules of chess from only standard algebraic chess notation (SAN) game data; 2) how much data is required to learn the rules well, and does learning scale proportionally with the data; and 3) how well do instruction fine-tuned language models play the game of chess?

\begin{figure}[ht]
\begin{center}
\includegraphics[width=120mm]{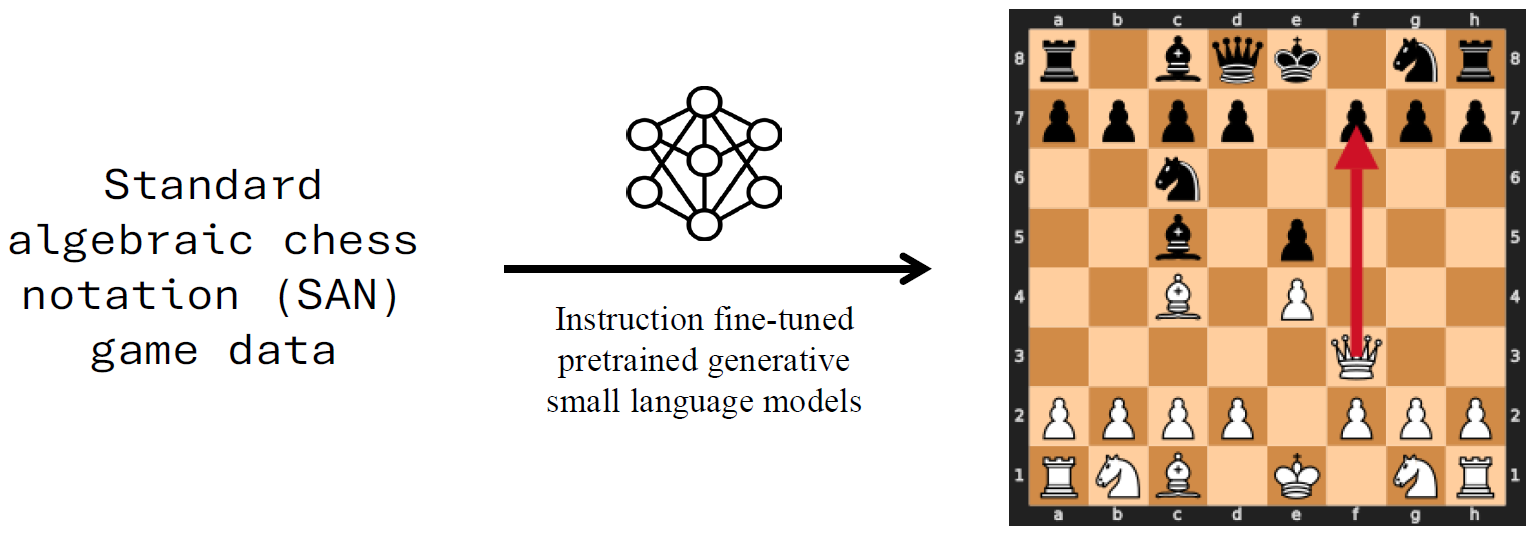}
\caption{Illustration of our proposed task: prediction of a white chess piece move (red arrow) given a board state in standard algebraic chess notation (SAN). The chessboard has standard algebraic notation ranks and files along the board edges.}
\label{overview}
\end{center}
\end{figure}

\section{Background}

Chess is a strategic board game played between two opponents on an $8 \times 8$ grid called a chessboard. Each player begins with 16 pieces of a single color, usually white or black: one king, one queen, two rooks, two knights, two bishops, and eight pawns. The player with the white pieces makes the first move to initiate the game. 

The objective is to checkmate the opponent’s king, meaning the king is in a position to be captured ("in check") and there is no legal move to escape the threat. Players take turns moving one piece at a time, with each type of piece having its own unique movement pattern. The game can also end in a draw under various conditions, such as a stalemate or insufficient material to achieve checkmate. Strategy and tactics are crucial, as players aim to control the board, capture opponent pieces, and protect their own king \cite{HarknessBlueBookChess}.

The game of chess is thought to of originated in India circa 600 CE. By 1000, the game had spread throughout Asia, the Middle East, and Europe.  The game was widely adopted by nobility in these regions around 1500, and thus became known as the "Royal Game" \cite{LombardyPanorama}.

Although the game was widely played circa 1500, the rules varied widely by region. The rules of the game were not unified and codified until 1851 during a European congress of players organized by St. George's Chess Club of London at the Crystal Palace \cite{LombardyPanorama}. The international chess code underwent some minor modifications in 1929 and 1954, but has since been largely unchanged \cite{HarknessBlueBookChess}.

Automated chess game play dates back to 1769 with the advent of the Mechanical Turk automaton by Wolfgang von Kempelen \cite{OxfordCompanion}. Serious interest in chess software began in the 1950s and gained momentum with the discovery and application of the alpha-beta pruning search algorithm to optimize move evaluation \cite{NewllShawSimonCMU}. A significant milestone was IBM's Deep Blue defeating then World Chess Champion Garry Kasparov in 1997, proving computers could compete at the highest levels \cite{HsuIBMDeepBlue}.\footnote{While Kasparov’s defeat by Deep Blue is perhaps the most famous instance of a computer beating a Grandmaster, Saviely G. Tartakower was actually the first Grandmaster to lose to a computer. This happened in 1951 when he was defeated by the \emph{Ajedrecista} chess computer in Paris.} Computers have since transformed chess by expanding opening theory, providing definitive endgame solutions, and bots/engines that are widely used for online play.\footnote{https://lichess.org/ (accessed 11Sept2024)} 

Deep Blue was an expert system that combined a vast database of chess knowledge and heuristics with a powerful tree-search algorithm. Most modern and stronger chess engines, such as Stockfish, follow a similar approach.\footnote{https://stockfishchess.org/ (accessed 11Sept2024)} Notable architectural exceptions include AlphaZero \cite{doi:10.1126/science.aar6404} and and its open-source counterpart, Leela Chess Zero.\footnote{https://github.com/LeelaChessZero/ (accessed 13Sept2024)} Both utilize a combination of deep neural networks (DNNs) trained on data from millions of chess games, and powerful search algorithms paired with reinforcement learning (RL) to select moves with the highest probability of winning.

A recent paper described the results of training a transformer-based model from scratch on a billions-scale move action-value annotated chess dataset \cite{ruoss2024grandmaster}. Unlike previous chess engines, this work did not explicitly use reinforcement learning (RL) or search algorithms. Instead, it heavily relied on the search capabilities of the Stockfish chess engine to accurately annotate billions of boards with move action-values, thereby teaching the new transformer model the best moves for a given board state to maximize the probability of winning.

\subsection{Our Contribution}

Dr. B., the fictitious character in Stefan Zweig's \emph{Schachnovelle}, led us to explore the real possibility of learning the latent rules of chess from game data. Dr. B. achieved Grandmaster-level chess skills by obsessively studying a book containing 150 championship chess games. In his book, each game was described solely in standard algebraic chess notation (SAN), with no accompanying board diagrams.

Chess engines, such as Stockfish, have learned game play from board states paired with powerful search algorithms to select the next best move \cite{OxfordCompanion}. AlphaZero paired that approach with reinforcement learning (RL) to emphasize and learn the strategies with the highest probabilities of winning. The recent approach of combining billions of Stockfish-annotated move action-values with supervised learning differs slightly from the chess engine and reinforcement learning (RL) methods \cite{ruoss2024grandmaster}. However, it still utilized billions-scale static move action-values derived from chess engines with the goal of creating a Grandmaster-level chess bot. Additionally, to our knowledge, all prior approaches to chess engines/bots have been built/trained from scratch with the sole purpose of playing chess.

Unlike these prior approaches, we used only SAN data from chess games and problems to replicate the situation experienced by Dr. B. in Zweig's novella. Further, we did not train our models from the ground-up, rather, we chose to explore the instruction fine-tuning of pretrained generative small language models (SLMs). We have previously demonstrated that instruction fine-tuning small pretrained generative language models leads to highly specialized models capable of accurately performing challenging tasks that the base models cannot perform \cite{Fauber2024AccuratePO, Fauber2024PretrainedGL}. In this work, we sought to address the following questions:

\begin{enumerate}
\item Can an instruction fine-tuned small generative language model learn the rules of chess from only standard algebraic chess notation (SAN) game data?
\item If a fine-tuned model can learn the rules, how much data is required to learn them well, and does learning scale proportionally with the data?
\item How well do instruction fine-tuned small language models play the game of chess?
\end{enumerate}

\section{Methods}

The python library \texttt{python-chess} provides extensive functionality to analyze portable game notation (PGN) format chess game files.\footnote{https://github.com/niklasf/python-chess (accessed 03Sept2024)} The library also contains functions to extract board status of a game before/after a move and to check the legality of proposed moves given a board status.

\subsection{Standard Algebraic Chess Notation}

Standard algebraic chess notation (SAN) assigns a unique two-character identifier to each of the 64 squares on a chessboard. The first character represents the file (column) of the square, labeled from "a" (leftmost or queenside) to "h" (rightmost or kingside). The second character represents the rank (row) of the square, numbered from "1" (bottom side, white's first rank) to "8" (top side, black's first rank). For example, the initial positions of some pieces are: white queen's rook at a1, white king at e1, black queen's knight pawn at b7, and black king's rook at h8 \cite{OxfordCompanion}.

SAN identifies each piece by a single letter. The standard English values include pawn = "P", rook = "R", knight = "N", bishop = "B", queen = "Q", and king = "K". In our studies, we used uppercase letters to represent white pieces and lowercase letters to represent black pieces.

\begin{figure}[ht]
\begin{center}
\includegraphics[width=120mm]{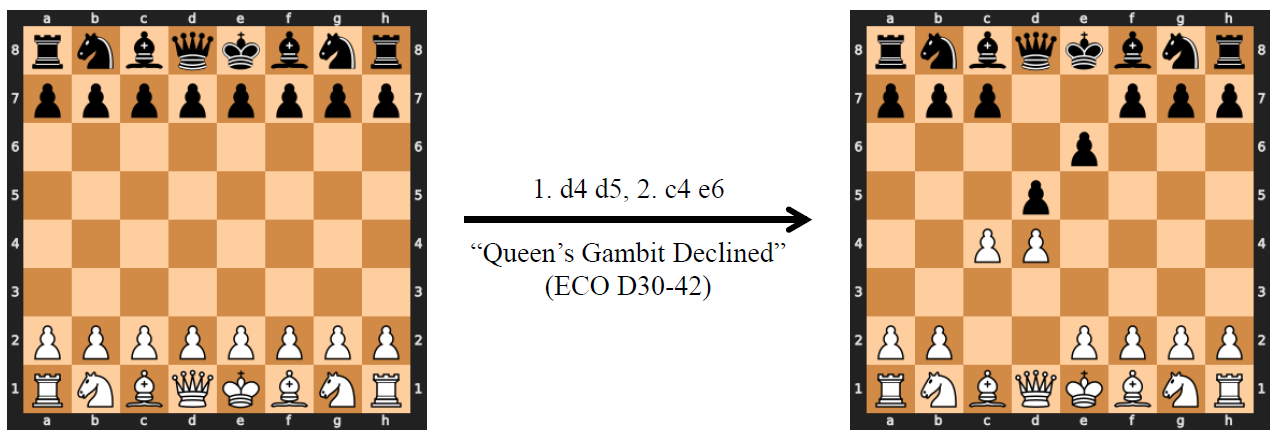}
\caption{Illustration of the initial array (left board diagram) and the standard algebraic chess notation (SAN) for the "Queen's Gambit Declined," a popular opening move sequence in the 1920's and 1930's (Encyclopedia of Chess Openings, sequences D30-42). The moves, in SAN, are as follows: 1. d4 d5, 2. c4 e6. The outcome of the two move sequences is shown on the right board diagram.}
\label{queens_gambit}
\end{center}
\end{figure}

Notably, the letter code for a pawn is not used in SAN moves within PGN format move text. However, letter codes for pawns and other pieces are sometimes necessary for certain tag pair and annotation constructs. For example, we used the letter codes for pawns in this study to differentiate between white and black pawn pieces when indicating our board status. An example of SAN move sequences is show in Figure 2, illustrating the "Queen's Gambit Declined" opening moves \cite{ECOLondon}.

A standard SAN move lists the moving piece’s letter (omitted for pawns) followed by the destination square. Capture moves are indicated by a lowercase "x" before the destination square. For pawn captures, the file letter of the capturing pawn's starting square is placed immediately before the "x".

In SAN notation, kingside castling is represented by the letters "O-O" and queenside castling by the letters "O-O-O". En passant captures are not specially notated; they are written as if the captured pawn were on the capturing pawn's destination square. Pawn promotions are indicated by an equal sign "=" immediately after the destination square, followed by the letter of the promoted piece (knight, bishop, rook, or queen) in uppercase.

If a move results in a check, a plus sign "+" is added to the end of the SAN notation. For a checkmate, a hash sign "\#" is used instead. There are no special notations for double checks, discovered checks, or drawing moves.

\subsection{Forsyth-Edwards Board Notation}

Forsyth-Edwards notation (FEN) denotes a single board status. It does not record the moves that lead to the board status. For example, the starting positions of the game pieces are denoted by the sequence "rnbqkbnr/pppppppp/8/8/8/8/PPPPPPPP/RNBQKBNR" \cite{StevenJEdwards}. Each row of the board is separated by a forward-slash "/" character, with uppercase letters representing white pieces and lowercase letters representing black pieces, using the same piece designations as in SAN.

\subsection{Dataset}

Stefan Zweig's novella \emph{Schachnovelle} was initially published in 1942 \cite{zweig_royalgame}. Therefore, we focused on games from prominent chess masters of that era as these were likely the games Dr. B., Zweig's character, might have studied. In the story, Dr. B. specifically mentions only two players: Alekhine and Bogoljubow, and their 1922 game in Pistyan, Slovakia.\footnote{This is different from the often-cited 1922 "Sacrificing the Queens" game between Alekhine and Bogoljubow at the Hastings Summer Chess Congress in England. In that game, Alekhine successively sacrificed three queens and was about to promote a pawn to a fourth queen when Bogoljubow resigned.}

The chess Grandmasters and International Chess Federation or FIDE (Fédération Internationale des Échecs) world champions leading up to the early 1940's included Alexander A. Alekhine (1892--1946) \cite{AlekhineMyBestVol1, AlekhineMyBestVol2, KotovAlekhine, BjelicaOnAlekhine}, José Raúl Capablanca (1888--1942) \cite{CapablancaPrimerOfChess, ReinfeldCapablanca, PanovCapablanca, ChernevCapablancasBest, BjelicaOnCapablanca, GolembekCapablancasBest}, Emanuel Lasker (1868--1941) \cite{LaskerManualOfChess, LaskerCommonSense, FineLaskersGreatest, CharushinLaskersCombo}, Aron I. Nimzowitsch (1886--1935) \cite{NimzowitschMySystem, NimzowitschPraxis}, Akiba K. Rubinstein (1880--1961) \cite{Kmoch100Rubenstein, RazuvaevRubinshtein}, and Savielly Tartakower (1887--1956) \cite{TartakowerHypermodern}. We included many game sources for these notable players. We also included texts for short games \cite{Chernev1000ShortGames}, openings \cite{MednisOpenings}, middle games \cite{LeiningerMiddlegame}, endings \cite{ChernevPracticalEndings}, and esteemed games \cite{Tartakower500Games, LombardyPanorama, ChernevMostInstructive, Chernev12GreatPlayersGames} to cover key tactics of game play. 

All chess game data was collected from the referenced texts as portable game notation (PGN) format files, a standard format for recording single or multiple games in a text file.\footnote{Games in PGN format were downloaded from: http://billwall.phpwebhosting.com/ (accessed 03Sept2024)} The PGN format contains the chess board configuration at the initiation of a game, or sequence, where the board state is encoded as a Forsyth-Edwards notation (FEN) text string \cite{StevenJEdwards}. Each white/black move pair are encoded within the PGN using standard algebraic chess notation (SAN). Illegal moves are not permitted in PGN move text.

All games were parsed into single moves for both white and black pieces, along with the board status before each move. The FEN game boards were converted into SAN, ensuring that both the move and board status were in SAN. The conversion to SAN was conducted to align with the experience of Stefan Zweig's character, Dr. B., in \emph{Schachnovelle}.

Parsing of the PGN files associated with all the above-cited publications resulted in a total of 20 million board and single-move combinations for white and black pieces. Filtering to only white piece single-moves resulted in 10 million board and move combinations. This data is referred to as the WSM-10M dataset. All moves were checked to ensure they were legal for the given board state. 

Each game in the WSM-10M dataset contained a mean of 39 white piece single-moves, with the longest game in the dataset containing 108 white piece single-moves. Further, there were 80k unique board and white piece single-move combinations in the dataset. This data is referred to as the Unique-WSM-10M dataset (\emph{see} the Appendix for additional details).

\subsection{Data Sampling}

Following best practices in machine learning, we randomly divided parent datasets into training/fine-tuning data and test data by sampling without replacement. We varied the number of available fine-tuning data instances from 1,000 to 1,000,000 examples of board states and their subsequent white piece single-moves, by random selection without replacement from the parent pool of fine-tuning data instances for each instance cohort. 

The fine-tuning data instances were used for language model instruction fine-tuning. The language models were never exposed to the test data (\emph{i.e.}, out-of-sample "hold-out" data) during the fine-tuning process to avoid train/test data contamination. 

We acknowledge that early-game board states are limited, and players frequently use popular openings. Consequently, these common board states and moves appear in both the fine-tuning and test datasets. We decided to keep them in both sets to avoid distributional shifts that could skew our evaluation metrics.

All data was formatted into an instruction-based format where the "instruction" was the input, and the "output" was the desired outcome. For example, the instruction was, "You are a chess Grandmaster and checkmate \# is your goal. Predict the next best move on this SAN chess board: h1:K, a2:P, g2:P, h3:P, b4:p, g4:R, f5:r, a6:R, f6:p, b7:r, f7:k," and the corresponding output was the white piece single-move, "Rg3". The instruction formatting was consistent throughout the fine-tuning and testing datasets unless otherwise noted.

\subsection{Pretrained Foundational Small Language Models}

We selected the OPT (open pretrained transformer) family of pretrained foundational generative language models as the starting point for our studies \cite{Zhang2022OPTOP}. We also explored the TinyStories \cite{Eldan2023TinyStoriesHS} family of language models. The OPT-125M model contained 125M parameters, whereas the TinyStories-28M model contained 28M parameters. Both models provided up to 2,048 positional embeddings for their inputs, permitting context for long string sequences which can be present in our method.

In our work, we defined model fine-tuning as initialization of a pretrained foundational language model followed by updates to the model weights and biases. In our fine-tuning setting, all language model parameters could undergo gradient updates -- there were no frozen layers nor adapters. In our prior work \cite{Fauber2024PretrainedGL}, we found the full fine-tuning approach was superior to adapter-based methods like LoRA (Low-Rank Adaptation) \cite{Hu2021LoRALA}. Other research groups have since confirmed our initial findings \cite{biderman2024lora}.

The prompt for the language models was consistent throughout our evaluation and across all models. The language model prompt was general and agnostic to the dataset instructions. The prompt used for our evaluation was: "Below is an instruction that describes a task. Write a response that appropriately completes the request. \#\#\# Instruction: \{instruction\} \#\#\# Response:".

\subsection{Evaluation of Our Method}

We evaluated the performance of our instruction fine-tuned language models on their ability to correctly propose legal white piece single-moves given a board state in the test dataset. We also evaluated the ability of our fine-tuned models to correctly predict the next white piece single-move in a "check or checkmate in one move" chess problem. The problems were drawn from a parent dataset of 1,836 chess problems \cite{PolgarChessProblems}. This data is referred to as the "Check/Mate-in-1" dataset.

We applied our previously-described instruction fine-tuning and text generation framework to ensure consistent outcomes and scoring \cite{Fauber2024PretrainedGL}. There were no detectable deviations in our study when replicate training sessions and fine-tuned SLM text generation results were evaluated.

\section{Results}

\subsection{Baseline Performance}

Language models can be generalists, or specialists, and it is valuable to understand what is required to create a specialist language model. It is important to determine how much fine-tuning data, and which fine-tuning paradigm, are reasonable starting points to create a specialist language model. 

We recognize that fine-tuning language models over multiple epochs may obliterate some portion of information that resides within the pretrained foundational language model. This potential change did not concern us as our objective was to create specialized language models from pretrained foundational language models, with the objective of effectively executing a highly specialized task that the original pretrained foundation models were incapable of performing. 

We evaluated a series of pretrained generative foundational language models on their ability to provide a legal chess move in SAN given a board state in SAN. Many popular pretrained foundational language models \cite{radford2019language, gpt-neo, Scao2022BLOOMA1, Zhang2022OPTOP, Eldan2023TinyStoriesHS, jiang2023mistral7b, dubey2024llama3herdmodels} were unable to perform this task with any reasonable level of proficiency (Table 1). 

\begin{table*}[ht]
\begin{center}
\begin{small}
\begin{tabular}{lccccc}
\toprule
\specialcellcenter{Pretrained Foundational \\ Language Model} & \specialcellcenter{Language Model \\ Parameters} & 
\specialcellcenter{\% Legal \\ Moves} \\
\midrule
roneneldan/TinyStories-28M & 28M & 0\% \\
gpt2 & 124M & 0\% \\
EleutherAI/gpt-neo-125m & 125M & 0\%	\\
facebook/opt-125m & 125M & 0\%	\\
facebook/opt-350m & 350M & 0\%	\\
facebook/opt-1.3b & 1.3B & 0\%	\\
facebook/opt-6.7b & 6.7B & 0\%	\\
mistralai/Mistral-7B-Instruct-v0.3 & 7B & 0\%	\\
bigscience/bloom-7b1 & 7.1B & 0\%	\\
meta-llama/Meta-Llama-3.1-8B-Instruct & 8B & 0\%	\\
\bottomrule
\end{tabular}
\end{small}
\caption{Baseline performance of pretrained foundational generative language models on their ability to provide a legal chess move in SAN for given a board state in SAN. Assessment used 1,000 test instances of board states in SAN drawn from the WSM-10M dataset as the model input. The language models are described by their \texttt{HuggingFace.co} repo names (accessed 13Sept2024)}
\end{center}
\end{table*}

\subsection{Increasing Dataset Size Improves Performance}

Our objective was to expose small pretrained foundational language models to increasing orders of magnitude of domain-specific instruction fine-tuning data and assess their performance against test data with well-defined metrics. In our primary assessment, we evaluated the percentage of legal proposed moves as a function of increasing orders of magnitude of instruction fine-tuning data. 

We observed that increasing the amount of domain-specific instruction fine-tuning data, as well as increasing the pretrained foundational language model parameter count (\emph{i.e.}, model size), improved the fine-tuned language model performance on our task (Figure 3). Thus, learning the latent rules of the game did scale with the data. Further, the OPT-125M model instruction fine-tuned with 1 million examples nearly always proposed a legal move.

\begin{figure}[ht]
\begin{center}
\includegraphics[width=100mm]{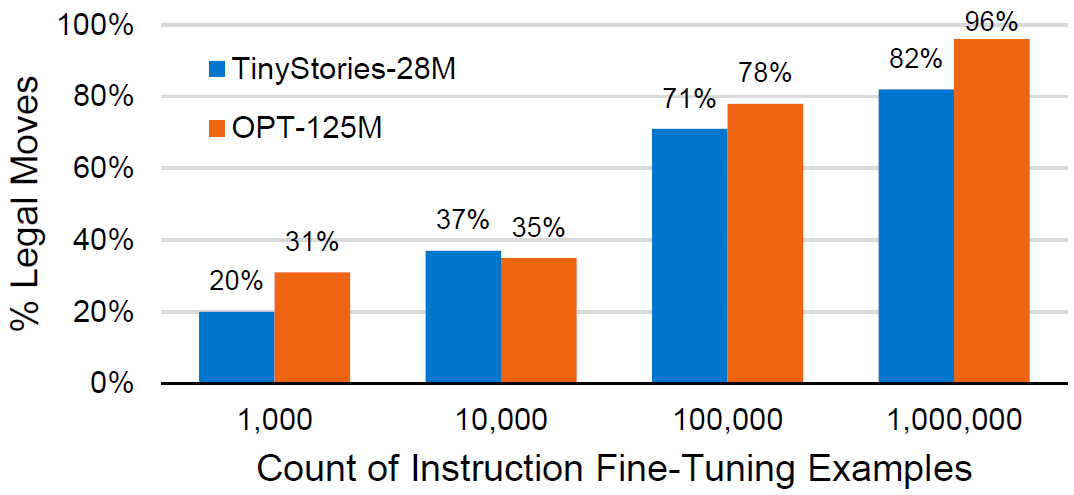}
\caption{Influence of increasing instruction fine-tuning examples. Percentage of legal proposed moves
versus count of the instruction fine-tuning examples for the TinyStories-28M (blue) and OPT-125M (orange) language models, instruction fine-tuned with learning rate = 2e-4, batch size = 4, and epochs = 3. The performance of the instruction fine-tuned language models was evaluated using 10,000 test instances of chess board states drawn from WSM-10M to assess the model's ability to generate a legal proposed move.
}
\label{scaling_legal}
\end{center}
\end{figure}

Next, we explored the influence of unique board and move combinations in the instruction fine-tuning dataset. Filtering the WSM-10M dataset to only unique board and move combinations resulted in the Unique-WSM-10M dataset, which contained approximately 80k white piece single-moves. The percentage of legal moves proposed by OPT-125M language models, which were instruction fine-tuned on progressively larger amounts of Unique-WSM-10M data, showed similar performance to the unfiltered WSM-10M fine-tuned models on the same number of example cohorts (Table 2). 

This result was notable because some examples in the WSM-10M dataset included duplicate board and move state combinations as players typically use similar opening moves and strategies. Therefore, filtering to only unique board and move state combinations did not provide beneficial outcomes when instruction fine-tuning for this task.  

\begin{table*}[ht]
\begin{center}
\begin{small}
\begin{tabular}{cccc}
\toprule
\specialcellcenter{Pretrained Foundational \\ Language Model} & \specialcellcenter{Instruction Fine-Tuning \\ Dataset} &
\specialcellcenter{Instruction Fine-Tuning \\ Example Count} &
\specialcellcenter{\% Legal \\ Moves} \\
\midrule
OPT-125M & Unique-WSM-10M & 1,000 & 29\%	\\
OPT-125M & Unique-WSM-10M & 10,000 & 36\%	\\
\bottomrule
\end{tabular}
\end{small}
\caption{Percentage of legal proposed moves versus count of the instruction fine-tuning examples for the OPT-125M language model, instruction fine-tuned with learning rate = 2e-4, batch size = 4, and epochs = 3. The instruction fine-tuning examples for these instances were drawn exclusively from the Unique-WSM-10M dataset. The Unique-WSM-10M dataset contained only 80k examples. Therefore, our assessment was limited to instruction fine-tuning cohorts of 1,000 and 10,000 examples, as larger datasets were not available for instruction fine-tuning. The performance of each instruction fine-tuned language model was evaluated using 10,000 test instances of chess board states drawn from WSM-10M to assess the model's ability to generate a legal proposed move.}
\end{center}
\end{table*}

\subsection{Chess Game Play Improves with More Data}

We evaluated the game play performance of our instruction fine-tuned models using check or checkmate in one white piece single-move chess problems \cite{PolgarChessProblems}. The data for this evaluation process was our Check/Mate-in-1 dataset. Our instruction fine-tuned models were evaluated on: 1) their abilities to both propose legal moves and; 2) their abilities to propose legal moves that resulted in either check or checkmate to solve the chess problems.

Our Check/Mate-in-1 dataset contained 1,836 examples. We sampled 1,000 test examples from this parent dataset for our evaluation. In our test instance containing 1,000 examples, the black king piece resided on 61 of the possible 64 squares of the board, representing a diversity of locations and strategies required to solve the problems. The test set problems contained between 3 and 32 pieces on the board. On average, each test set problem included a total of seven pawns, two knights, two bishops, three rooks, two queens, and two kings. 

\begin{figure}[ht]
\begin{center}
\includegraphics[width=100mm]{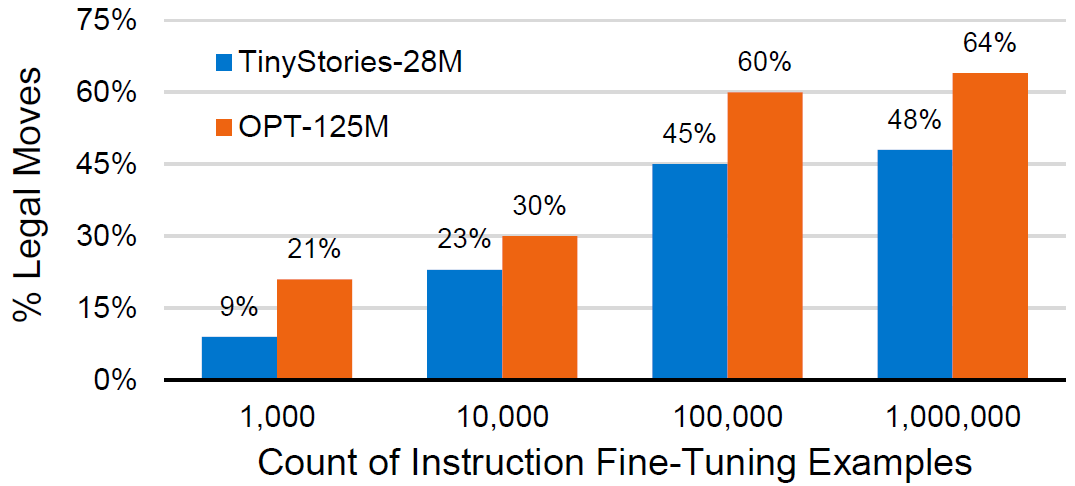}
\caption{Percentage of legal proposed moves versus count of the instruction fine-tuning examples for the TinyStories-28M (blue) and OPT-125M (orange) language models, instruction fine-tuned with learning rate = 2e-4, batch size = 4, and epochs = 3. The performance of each instruction fine-tuned language model was evaluated using 1,000 test instances of chess problems drawn from Check/Mate-in-1 to assess the model's ability to generate a legal proposed move.
}
\label{scaling_puzzle_legal}
\end{center}
\end{figure}

Both the TinyStories-28M and OPT-125M instruction fine-tuned models demonstrated reasonable abilities to propose legal moves (Figure 4) and solve the chess problems with legal moves (Figure 5). Again, the instruction fine-tuned OPT-125M models outperformed the smaller TinyStories-28M models. As noted in our earlier studies, the ability to accurately perform both tasks improved as the number of instruction fine-tuning examples increased.

\begin{figure}[ht]
\begin{center}
\includegraphics[width=100mm]{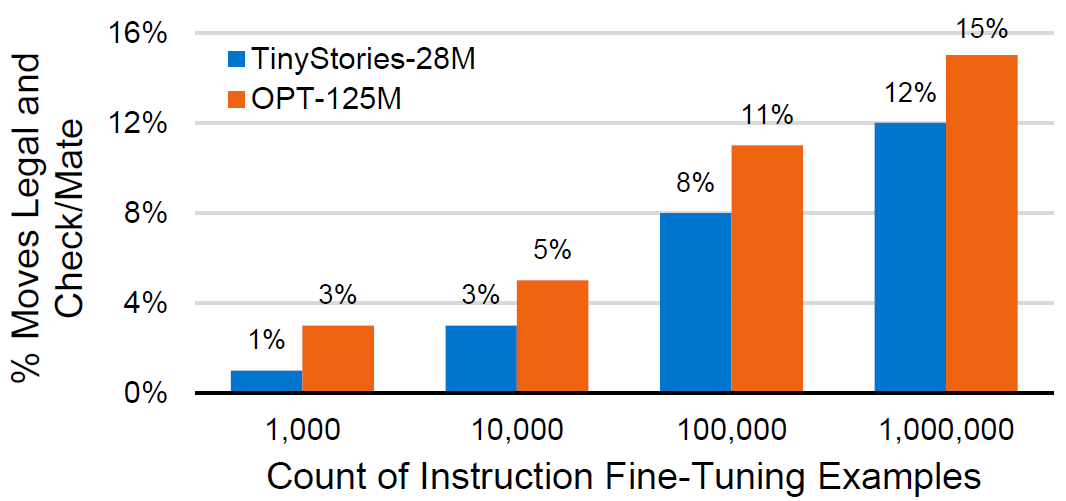}
\caption{Percentage of proposed moves which were legal and resulted in check or checkmate versus count of the instruction fine-tuning examples for the TinyStories-28M (blue) and OPT-125M (orange) language models, instruction fine-tuned with learning rate = 2e-4, batch size = 4, and epochs = 3. The performance of each instruction fine-tuned language model was evaluated using 1,000 test instances of chess problems drawn from Check/Mate-in-1 to assess the model's ability to generate a legal move that resulted in check or checkmate.}
\label{scaling_puzzle_check}
\end{center}
\end{figure}

\subsection{Hallucinations Decrease with More Instruction Fine-Tuning Examples}

While evaluating the performance of our instruction fine-tuned SLMs and their abilities to propose legal moves that resulted in check or checkmate on the Check/Mate-in-1 dataset, we also observed interesting trends regarding hallucinations. Generative language models are trained to accurately predict the next token in a sequence, thus generating human like text that resembles their training data. Yet, it is well known that generative language models can create text that sometimes contain inaccuracies or nonsensical information known as "hallucinations" \cite{agrawal2024languagemodelsknowtheyre}.

We noted that increasing orders of magnitude of instruction fine-tuning data resulted in fine-tuned OPT-125M models which proposed fewer illegal moves and hallucinations of pieces on a board which did not exist (Figure 6). Based on these results, we concluded that the instruction fine-tuned language models with fewer examples exhibited game play somewhat akin to a novice child. 

Typically, when teaching children to play chess, they quickly grasp the objective and may suggest illegal moves to rapidly achieve check or checkmate. We observed a similar outcome with the 1,000 and 10,000 example instruction fine-tuned OPT-125M models, where the models suggested a majority of illegal moves to achieve check or checkmate. Yet, these tendencies diminished as the number of instruction fine-tuning examples increased. 

Further, we observed that the tendency to hallucinate a piece onto a board to achieve check or checkmate was ablated as the number of instruction fine-tuning examples increased (Figure 6). The proposal of illegal moves to achieve check or checkmate decreased as the number of instruction fine-tuning examples were increased but was not totally eliminated with even 1 million instruction fine-tuning examples. Additionally, the percentage of illegal proposed moves to achieve check or checkmate (Figure 6) was much higher compared to legal moves proposed for the same goals (Figure 5). Therefore, we concluded that understanding the objective of the game was easier than learning its rules.

\begin{figure}[ht]
\begin{center}
\includegraphics[width=100mm]{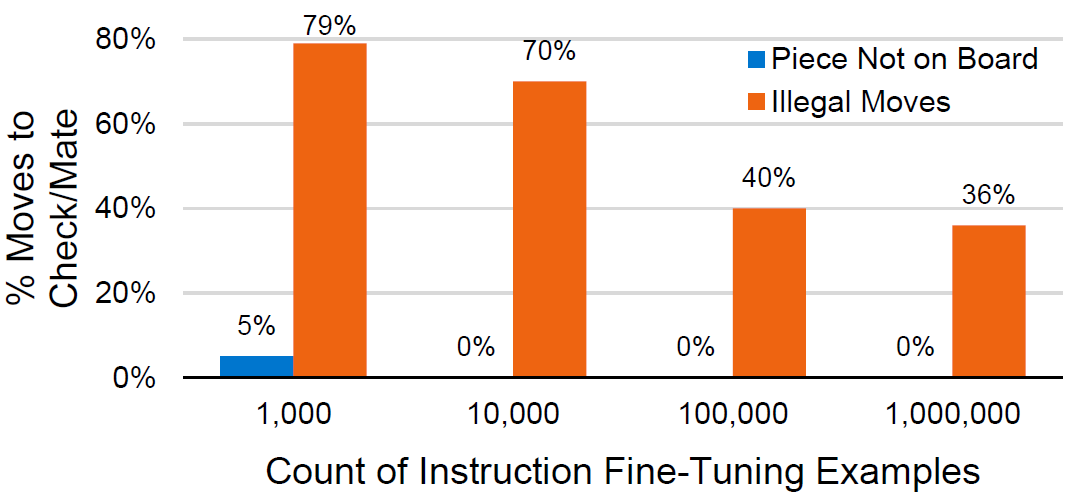}
\caption{Percentage of proposed moves which were either illegal (orange), or the proposed piece was not on the board (blue), to achieve either check or checkmate versus count of the instruction fine-tuning examples for the OPT-125M language model, instruction fine-tuned with learning rate = 2e-4, batch size = 4, and epochs = 3. The performance of each instruction fine-tuned language model was evaluated using 1,000 test instances of chess problems drawn from Check/Mate-in-1 to assess the model's propensity to generate an illegal move that resulted in check or checkmate.}
\label{scaling_puzzle_hallucinate}
\end{center}
\end{figure}

\subsection{Models Learn the Game Objective from Game Data}

We also explored the role of the instruction fine-tuning text in achieving a legal move which resulted in check or checkmate. For all instruction fine-tuning examples, we included the statement, "You are a chess Grandmaster and checkmate \# is your goal." One could posit that this statement explicitly states the game's objective, and the model was learning this objective from the instruction text, along with all moves marked with the hash sign "\#" to indicate checkmate. To evaluate the role of this statement in the instruction fine-tuning text, we used the exact same 1,000 to 1,000,000 instruction fine-tuning cohorts for WSM-10M \emph{without} the statement, "You are a chess Grandmaster and checkmate \# is your goal." We referred to this revised instruction fine-tuning dataset as NoGoal-WSM-10M. 

\begin{table*}[ht]
\begin{center}
\begin{small}
\begin{tabular}{ccccc}
\toprule
\specialcellcenter{Pretrained Foundational \\ Language Model} & \specialcellcenter{Instruction Fine-Tuning \\ Dataset} &
\specialcellcenter{Instruction Fine-Tuning \\ Example Count} &
\specialcellcenter{\% Legal \\ Moves} &
\specialcellcenter{\% Legal and \\ Check/Mate Moves} \\
\midrule
OPT-125M & NoGoal-WSM-10M & 1,000 & 16\% & 1\%	\\
OPT-125M & NoGoal-WSM-10M & 10,000 & 33\% & 6\%	\\
OPT-125M & NoGoal-WSM-10M & 100,000 & 52\% & 10\%	\\
OPT-125M & NoGoal-WSM-10M & 1,000,000 & 60\% & 12\%	\\
\bottomrule
\end{tabular}
\end{small}
\caption{Percentage of proposed moves which were legal and resulted in check or checkmate versus count of the instruction fine-tuning examples for the OPT-125M language model, instruction fine-tuned with learning rate = 2e-4, batch size = 4, and epochs = 3. The instruction fine-tuning examples for these instances were drawn exclusively from the NoGoal-WSM-10M dataset. The performance of each instruction fine-tuned language model was evaluated using 1,000 test instances of chess problems drawn from Check/Mate-in-1 to assess the model's ability to generate a legal move and a legal move that resulted in check or checkmate.}
\end{center}
\end{table*}

We found that usage of the WSM-10M cohort or the NoGoal-WSM-10M cohort for instruction fine-tuning the OPT-125M model resulted in essentially the same outcomes for the percentage of legal proposed moves and legal proposed moves which resulted in check or checkmate (Table 3, Figure 4, and Figure 5). The language models fine-tuned with the NoGoal-WSM-10M datasets also showed similar results with regards to piece hallucinations and illegal moves proposed to achieve check or checkmate, as observed with the WSM-10M fine-tuned models (\emph{see} Appendix for additional details). Therefore, we concluded that the instruction text was not teaching the model the game objective. Instead, the model was learning the game objective from the game play data.

\subsection{Revisiting the Same Data Provides Limited Benefits}

We examined the model's ability to learn from a dataset by repeatedly revisiting the data and then evaluated its performance after multiple revisits. This process was analogous to Stefan Zweig's character Dr. B. obsessively studying his book of championship chess games.

We discovered that increasing the number of instruction fine-tuning epochs for the OPT-125M language model initially improved the model's ability to perform our tasks (Figure 7). This outcome was consistent for both the percentage of legal proposed moves and the legal moves that resulted in check or checkmate on the Check/Mate-in-1 dataset.

These results also demonstrated the potential to improve an instruction fine-tuned model's performance for a task with a constrained amount of instruction fine-tuning data. Yet, the successive improvements in results from increasing epochs do appear to plateau after 5-10 fine-tuning epochs.

\begin{figure}[ht]
\begin{center}
\includegraphics[width=100mm]{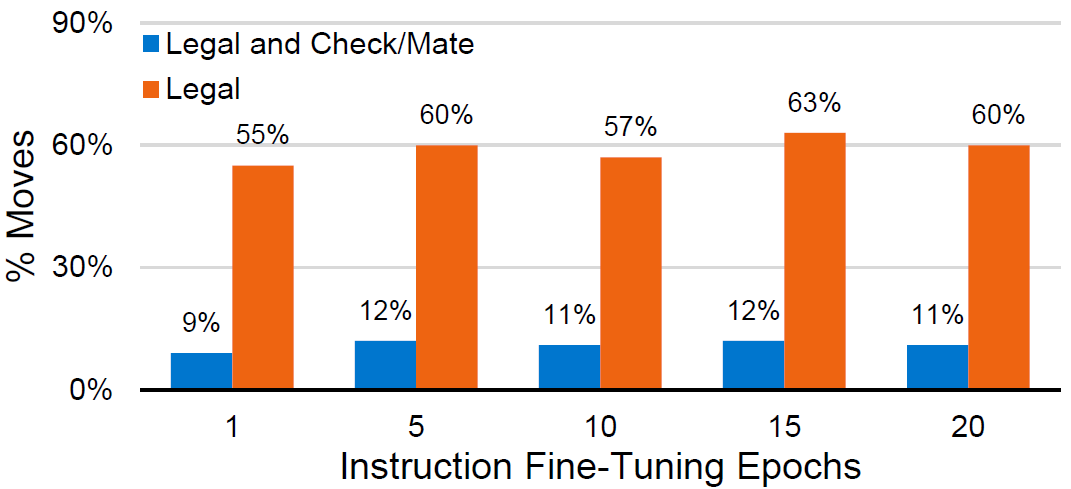}
\caption{Percentage of proposed moves which were legal (orange) and resulted in check or checkmate (blue) versus count of the instruction fine-tuning epochs for the OPT-125M language model, instruction fine-tuned with 100,000 examples drawn from the WSM-10M dataset, learning rate = 2e-4, and batch size = 4. The performance of each instruction fine-tuned language model was evaluated using 1,000 test instances of chess problems drawn from Check/Mate-in-1 to assess the model's ability to generate a legal move that resulted in check or checkmate.}
\label{scaling_puzzle_epochs}
\end{center}
\end{figure}

\subsection{Requiring Legal Moves with Increased Temperature Improves Outcomes}

Throughout our studies, we used the default settings for the \texttt{transformers.GenerationConfig()} text generation function, unless otherwise specified (\emph{see} Appendix). We also disabled the \texttt{do\_sample} hyperparameter within that function to ensure consistent and reproducible text generation, as documented in our previous work \cite{Fauber2024AccuratePO, Fauber2024PretrainedGL}.

Conversely, enabling of the \texttt{do\_sample} hyperparameter makes use of the temperature hyperparameter to allow for variation in text generation based on the temperature setting (default = 1.0). Thus, we explored the role of increasing generation variability, with the temperature hyperparameter $\ge 1.0$, and a requirement that the proposed moves be legal. 

In this analysis, the text generation process was allowed up to 100 iterations to obtain a legal proposed move. If no legal move was proposed by the final iteration, the attempt was labeled as illegal, and the process moved on to the next test example. For this study, we used an OPT-125M language model that was instruction fine-tuned with 1,000,000 examples drawn from the WSM-10M dataset.

The baseline text generation for this study, where the \texttt{do\_sample} hyperparameter was disabled (temperature = 1.0 as the default setting) and only a single text generation iteration was allowed, is shown in Figures 4 and 5. Conversely, enabling the \texttt{do\_sample} hyperparameter and allowing up to 100 text generation iterations to achieve a legal proposed move, while keeping the temperature at the default setting of 1.0, resulted in a higher percentage of proposed legal moves and legal moves that led to check or checkmate (Figure 8).

Increasing the temperature hyperparameter under the same text generation conditions provided further improvements in proposed legal moves and legal moves that resulted in check or checkmate, until the process plateaued at temperature = 3.5 (Figure 8). At this point, $>99\%$ of the proposed moves were legal, and $24\%$ of the proposed moves were legal and resulted in check or checkmate. Further increasing the text generation temperature did not improve the percentage of legal moves, and only resulted in some small aberrations in the percentage of legal moves that resulted in check or checkmate. 

These results showed that increasing text generation variability to allow the model to be more creative, while ensuring the outputs remained legal, improved the results, but there were limits. This study also revealed that the instruction fine-tuned model contained some additional knowledge about proposing a legal move for check or checkmate given a board state, which was only disclosed when the model was forced to generate a legal move.

\begin{figure}[ht]
\begin{center}
\includegraphics[width=100mm]{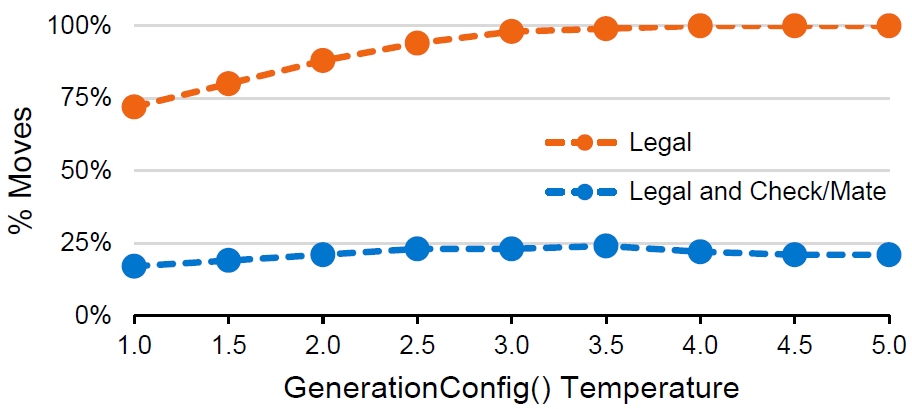}
\caption{Percentage of proposed moves which were legal (orange) and resulted in check or checkmate (blue) versus the \texttt{GenerationConfig()} temperature parameter value when required to generate a legal move based on the board state. The model was allowed up to 100 iterations to achieve this requirement before proceeding to the next example in the test dataset. All instances used an instruction fine-tuned OPT-125M language model, instruction fine-tuned with 1,000,000 examples drawn from the WSM-10M dataset, learning rate = 2e-4, and batch size = 4. The performance of each instruction fine-tuned language model was evaluated using 1,000 test instances of chess problems drawn from Check/Mate-in-1 to assess the model's ability to generate a legal move that resulted in check or checkmate.}
\label{legal_moves_by_temp}
\end{center}
\end{figure}

\section{Discussion}

We have provided a framework for rigorous and systematic evaluation of our instruction fine-tuned language model outputs. We demonstrated that small pretrained foundational generative language models with millions of parameters can learn the latent rules of a process from data associated with the process. 

We were inspired by Stefan Zweig's novella \emph{Schachnovelle}, also known as \emph{The Royal Game} in English, in which his character, Dr. B., learns how to play Grandmaster-level chess by compulsively studying a book of 150 championship chess games. We showed that 28M and 125M parameter pretrained foundational language models can be instruction fine-tuned with 1,000-to-1,000,000 instruction examples to learn the rules of chess, propose legal moves, and accurately solve chess problems. 

We also demonstrated that increasing the number of instruction fine-tuning examples leads to models learning the objective of the game before learning the rules of the game. We also showed that increasing the number of instruction fine-tuning examples minimizes hallucinations associated with illegal proposed moves and imagining pieces on the game board to enable check or checkmate.

Finally, we showed that increasing the number of instruction fine-tuning epochs can lead to some improvements in legal moves and accurately solve chess problems, but the improvements quickly plateau after 5-10 epochs. At a high level, these results further demonstrate that pretrained generative language models can serve as general learning frameworks for sequence-based tasks.

Overall, our results suggest that instruction fine-tuning generative language models with chess game play data can replicate the learning experience of Stefan Zweig's character, Dr. B, in \emph{Schachnovelle}. Specifically, increasing the amount of SAN-formatted game data used for instruction fine-tuning improves our models' understanding of the game rules and game play. 

However, the amount of data Zweig's character encountered in his book of 150 championship chess games was quite small, with an estimated 40 white piece single moves per game, totaling around 6,000 white piece single moves in his book. In contrast, our systems required at least 1,000,000 white piece single-move examples to successfully learn the objectives and rules of chess. Even then, instruction fine-tuning models with 1,000,000 examples was insufficient to solve all the chess problems in our Check/Mate-in-1 dataset. This is certainly a lesser outcome compared to the fictitious, yet plausible, character Dr. B in \emph{Schachnovelle}. Overall, these results suggest that while small generative language models can learn the latent rules of a game from game play data alone, they need significantly more data than humans to do so.

\section{Conclusion}

In conclusion, our results indicate that instruction fine-tuned generative language models can learn the rules of chess from only standard algebraic chess notation (SAN) game data. Further, model performance scales increasingly well with increasing orders of magnitude of instruction fine-tuning data. For example, instruction fine-tuning of the OPT-125M generative language model on 1 million examples resulted in nearly perfect proposals of legal moves when evaluated on 10,000 test examples drawn our WSM-10M dataset. Finally, we demonstrated that our instruction fine-tuned language models can propose winning strategies for chess problems.

\section{Acknowledgements}

The author would like to thank Anas Bricha for supporting this project and Guy Laporte for providing access to the computational infrastructure to conduct these studies. The author declares no financial interests nor conflicts. 

\newpage

\bibliography{bib-llms}
\bibliographystyle{icml2020}

\newpage

\appendix

\setcounter{table}{0}
\renewcommand{\thetable}{A\arabic{table}}

\setcounter{figure}{0}
\renewcommand{\thefigure}{A\arabic{figure}}

\section{Appendix}
\subsection{Computational Infrastructure and Code}

The results described in this article were carried out using a Dell Technologies PowerEdge C4140 server with 4 x V100 NVIDIA\textsuperscript{\textregistered} SXM GPU cards with 32 GB VRAM each and NVLink\textsuperscript{TM} connectivity. There were 2 x Intel\textsuperscript{\textregistered} Xeon\textsuperscript{\textregistered} processors on the server with 1.5 TB of CPU RAM. 

The server was configured with the Ubuntu v22.04 Linux operating system, Anaconda v23.1.0, NVIDIA\textsuperscript{\textregistered} CUDA v12.2, and NVIDIA\textsuperscript{\textregistered} drivers v535.54.03. Additional python dependencies included: \texttt{python-chess  v1.10.0},
\texttt{torch v2.1.1}, and \texttt{transformers v4.45.1}.

The Stanford ALPACA language model code was git cloned directly from \url{https://github.com/tatsu-lab/stanford_alpaca} (accessed 30Dec2023). The \texttt{train.py} file in the GitHub repo, along with our corresponding instruction fine-tuning dataset, was used to instruction fine-tune the language models in our study. The language model fine-tuning code was executed via the command line interface (CLI).

As an example, the following CLI command was used to instruction fine-tune a pretrained foundational language model on 4 GPUs:

\begin{lstlisting}
torchrun --nproc_per_node=4 [TRAINING_PY_FILE] \
    --model_name_or_path [HUGGINGFACE_MODEL_NAME] \
    --data_path [DATA_PATH_TO_FORMATTED_JSON_FILE] \
    --bf16 False \
    --output_dir [OUTPUT_DIRECTORY] \
    --overwrite_output_dir True \
    --num_train_epochs 3 \
    --per_device_train_batch_size 4 \
    --per_device_eval_batch_size 4 \
    --gradient_accumulation_steps 8 \
    --save_strategy "steps" \
    --save_steps 5000 \
    --save_total_limit 1 \
    --learning_rate 2e-4 \
    --weight_decay 0. \
    --warmup_ratio 0.03 \
    --lr_scheduler_type "cosine" \
    --seed 41 \
    --logging_steps 1 \
    --tf32 False
\end{lstlisting}

\newpage

\subsection{Dataset Creation}

The dataset was focused on, but not exclusive to, games from prominent chess masters in the early 1900's to 1940's. The chess Grandmasters and International Chess Federation or FIDE (Fédération Internationale des Échecs) world champions leading up to the early 1940's included Alexander A. Alekhine (1892--1946) \cite{AlekhineMyBestVol1, AlekhineMyBestVol2, KotovAlekhine, BjelicaOnAlekhine}, José Raúl Capablanca (1888--1942) \cite{CapablancaPrimerOfChess, ReinfeldCapablanca, PanovCapablanca, ChernevCapablancasBest, BjelicaOnCapablanca, GolembekCapablancasBest}, Emanuel Lasker (1868--1941) \cite{LaskerManualOfChess, LaskerCommonSense, FineLaskersGreatest, CharushinLaskersCombo}, Aron I. Nimzowitsch (1886--1935) \cite{NimzowitschMySystem, NimzowitschPraxis}, Akiba K. Rubinstein (1880--1961) \cite{Kmoch100Rubenstein, RazuvaevRubinshtein}, and Savielly Tartakower (1887--1956) \cite{TartakowerHypermodern}. In addition to the above-cited texts, we also included texts for short games \cite{Chernev1000ShortGames}, openings \cite{MednisOpenings}, middle games \cite{LeiningerMiddlegame}, endings \cite{ChernevPracticalEndings}, and esteemed games \cite{Tartakower500Games, LombardyPanorama, ChernevMostInstructive, Chernev12GreatPlayersGames} to cover key tactics of game play. 

All chess game data was collected from the referenced texts as portable game notation (PGN) format files. All games were parsed into individual moves for both white and black pieces, along with the board status before each move. The FEN game boards were converted into SAN, ensuring that both the move and board status were in SAN. The conversion to SAN was conducted to align with the experience of Stefan Zweig's character, Dr. B., in \emph{Schachnovelle}.

\begin{figure}[ht]
\begin{center}
\includegraphics[width=80mm]{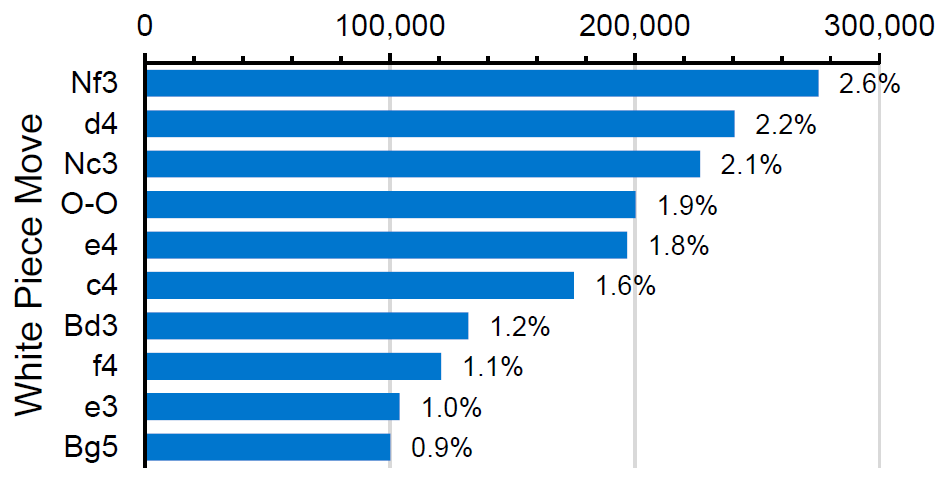}
\caption{Top 10 most prevalent white piece single-moves. Moves are shown by their standard algebraic chess notations (SAN). The $x$-axis of the plot displays the total single-move count for each move, with data labels indicating their percentage of the overall total.}
\label{top_10_white_piece_moves}
\end{center}
\end{figure}

Parsing of the PGN files associated with all the above-cited publications resulted in a total of 20M board and single-move combinations for white and black pieces. Filtering to only white piece single-moves resulted in 10M board and move combinations. The top 10 most prevalent white piece single-moves and proceeding board configurations are shown in Figures A1 and A2, respectively. Board diagrams of the top 8 (excluding the initial array) most common board configurations, and their associated Encyclopedia of Chess Openings (ECO) traditional names and codes \cite{ECOLondon}, and paired-move sequences, are shown in Figure A3. The notable chess players associated with the single-move data are shown in Figure A4. 

Each game in the WSM-10M dataset contained a mean of 39 white piece single-moves, with the longest game in the dataset containing 108 white piece single-moves. Further, there were 80k unique board and white piece single-move combinations in the dataset.

\newpage

\begin{figure}[ht]
\begin{center}
\includegraphics[width=100mm]{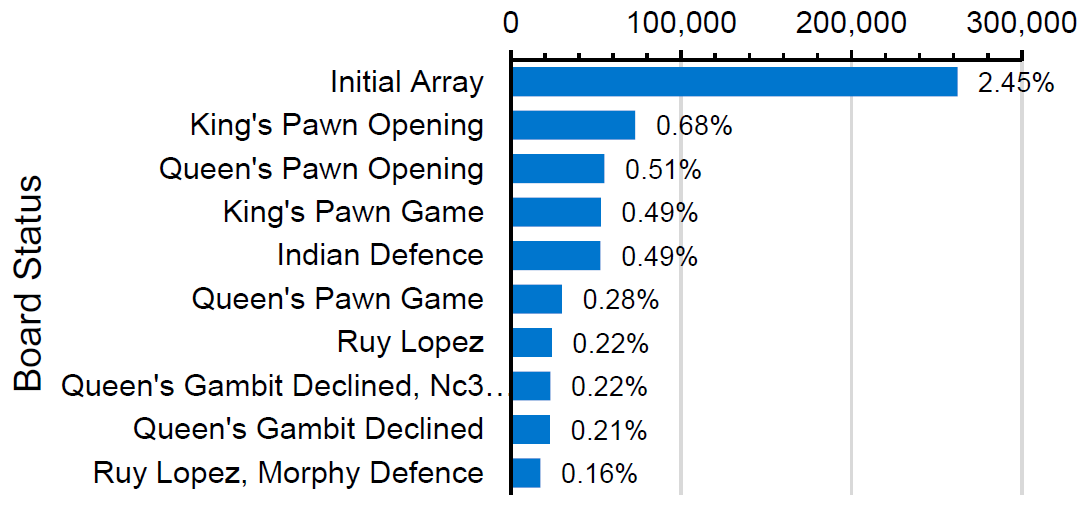}
\caption{Top 10 most prevalent chess boards preceding white piece single-moves (\emph{i.e.}, next move = white). Common move openings and sequences are described by their Encyclopedia of Chess Openings (ECO) traditional names. The $x$-axis of the plot displays the total board count for each board status, with data labels indicating their percentage of the overall total.}
\label{top_10_boards}
\end{center}
\end{figure}

\begin{figure}[ht]
\begin{center}
\includegraphics[width=150mm]{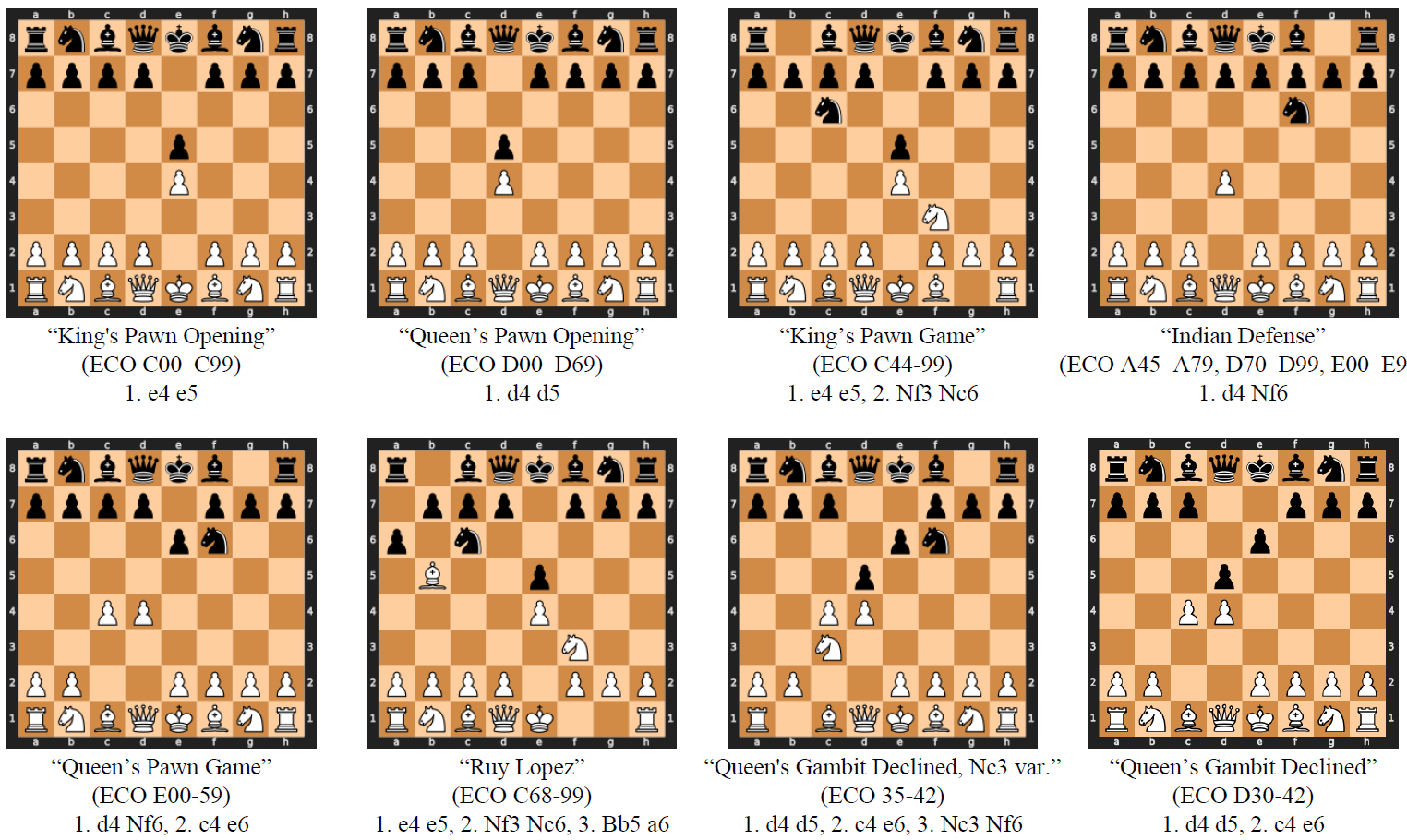}
\caption{Top 8 (excluding the initial array) most common board configurations in the white piece single-move dataset where white moves next. Their associated Encyclopedia of Chess Openings (ECO) traditional names and codes, and paired-move sequences in SAN, are included.}
\label{diagrams_top8_boards}
\end{center}
\end{figure}

\newpage

\begin{figure}[ht]
\begin{center}
\includegraphics[width=100mm]{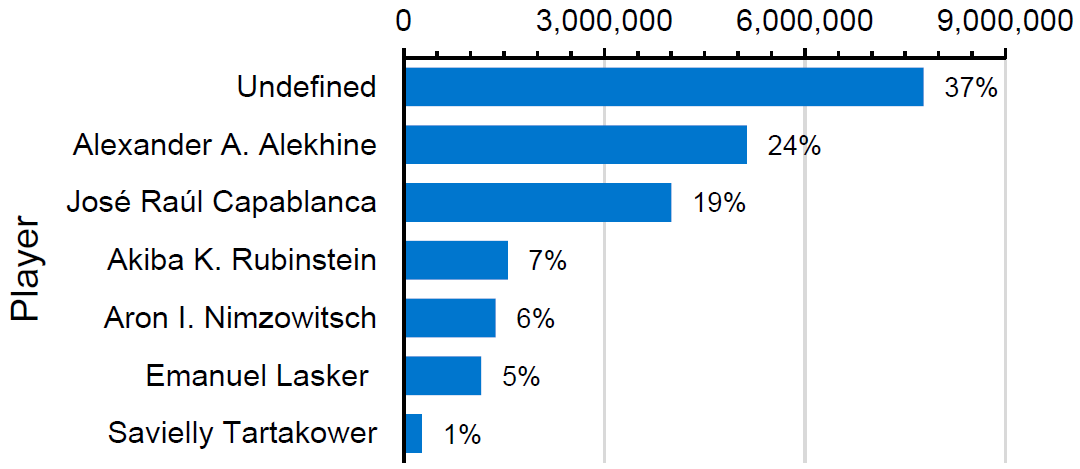}
\caption{Notable chess players in the dataset were identified by the main player linked to each book or data source, covering all games/setups and the associated 20 million single piece moves. The $x$-axis of the plot displays the single-move count for each player, with data labels indicating their percentage of the overall total.}
\label{sources_by_players}
\end{center}
\end{figure}

\newpage

\subsection{Language Model Text Generation Configuration}

The same language model fine-tuning and generation configurations were utilized throughout our studies, and only single-parameter changes were permitted, as annotated in the tables, when comparing methods. Language model text generation was conducted via the HuggingFace \texttt{transformers} library. Transformers \texttt{GenerationConfig()} was set to the default parameters, along with:
\begin{itemize}
    \item \texttt{num\_beams = 2}, 
    \item \texttt{repetition\_penalty = 1.3}, 
    \item \texttt{do\_sample = False} (for consistent output generation), 
    \item \texttt{early\_stopping = True}, 
    \item \texttt{max\_time = 10}, and 
    \item \texttt{length\_penalty = 0.4}.
\end{itemize}

In prior studies, we found the above configuration parameters provided stable and reproducible text generation \cite{Fauber2024PretrainedGL}. The text generation prompt and the general prompt used in the language model fine-tuning process were identical: 
\\
"\texttt{Below is an instruction that describes a task. Write a response that appropriately completes the request. \#\#\# Instruction: \{instruction\} \#\#\# Response: \{output\}}". 

Although not always necessary, we enforced truncation of the output text for all models to ensure consistency in outcomes. Truncation of the OPT model text output returned all text following the "\texttt{\#\#\# Response:}" string. Similarly, the TinyStories families of models truncated the output to the text following the "\texttt{<|endoftext|>}" string.

\newpage

\subsection{Analysis of the Instruction Text}

We explored the role of the instruction fine-tuning text in achieving a legal move which resulted in check or checkmate. For all instruction fine-tuning examples, we included the statement, "You are a chess Grandmaster and checkmate \# is your goal." One could posit that this statement explicitly states the game's objective, and the model was learning this objective from the instruction text, along with all moves marked with the hash sign "\#" to indicate checkmate. 

To evaluate the role of this statement in the instruction fine-tuning text, we used the exact same 1,000 to 1,000,000 instruction fine-tuning cohorts for WSM-10M \emph{without} the statement, "You are a chess Grandmaster and checkmate \# is your goal." We referred to this revised instruction fine-tuning dataset as NoGoal-WSM-10M. 

We found that usage of the WSM-10M cohort or the NoGoal-WSM-10M cohort for instruction fine-tuning the OPT-125M model resulted in essentially the same outcomes for the percentage of legal proposed moves and legal proposed moves which resulted in check or checkmate (Table 3, Figure 4, and Figure 5). The language models fine-tuned with the NoGoal-WSM-10M datasets also showed similar results with regards to piece hallucinations and illegal moves proposed to achieve check or checkmate, as observed with the WSM-10M fine-tuned models (Table A1 and Figure 6).

\begin{table*}[ht]
\begin{center}
\begin{small}
\begin{tabular}{ccccc}
\toprule
\specialcellcenter{Instruction Fine-Tuning \\ Dataset} &
\specialcellcenter{Instruction Fine-Tuning \\ Example Count} &
\specialcellcenter{\% Illegal and \\ Check/Mate Moves} &
\specialcellcenter{\% Piece Not on Board \\  and Check/Mate Moves} \\
\midrule
NoGoal-WSM-10M & 1,000 & 84\% & 3\%	\\
NoGoal-WSM-10M & 10,000 & 67\% & 0\%	\\
NoGoal-WSM-10M & 100,000 & 47\% & 0\%	\\
NoGoal-WSM-10M & 1,000,000 & 40\% & 0\%	\\
\bottomrule
\end{tabular}
\end{small}
\caption{Percentage of proposed moves that were either illegal, or the proposed piece was not on the board, to achieve either check or checkmate versus count of the instruction fine-tuning examples for the OPT-125M language model, instruction fine-tuned with learning rate = 2e-4, batch size = 4, and epochs = 3. The instruction fine-tuning examples for these instances were drawn exclusively from the NoGoal-WSM-10M dataset. The performance of each instruction fine-tuned language model was evaluated using 1,000 test instances of chess problems drawn from Check/Mate-in-1 to assess the model's propensity to generate an illegal move that resulted in check or checkmate.}
\end{center}
\end{table*}

\end{document}